\documentclass[letterpaper]{article}
\usepackage{CJKutf8}
\usepackage{aaai}
\usepackage{times}
\usepackage{helvet}
\usepackage{courier}
\usepackage{booktabs}
\usepackage{graphicx}
\usepackage{subcaption}
\usepackage{amsmath}
\usepackage{xcolor}
\usepackage{latexsym}
\usepackage{natbib}
\usepackage{tikz}
\usepackage{etoolbox}
\usepackage{booktabs}
\usepackage{tikzsymbols}
\usepackage{adjustbox}
\usepackage{verbatim}
\usepackage{caption}
\usepackage{multirow}
\begin{document}

\title{TopicRefine: Joint Topic Prediction and Dialogue Response Generation for Multi-turn End-to-End Dialogue System}

\author{ Hongru Wang\footnotemark[1], 
        Mingyu Cui\footnotemark[1], 
        Zimo Zhou, 
        Gabriel Pui Cheong Fung, 
        Kam-Fai Wong \\
    Department of System Engineering and Engineering Management \\
    The Chinese University of Hong Kong \\
   \{hrwang, kfwong\}@se.cuhk.edu.hk \\
}
\maketitle

\footnotetext[1]{Equal contributions.}

\begin{abstract}
A multi-turn dialogue always follows a specific topic thread, and topic shift at the discourse level occurs naturally as the conversation progresses, necessitating the model's ability to capture different topics and generate topic-aware responses. Previous research has either predicted the topic first and then generated the relevant response, or simply applied the attention mechanism to all topics, ignoring the joint distribution of the topic prediction and response generation models and resulting in uncontrollable and unrelated responses. In this paper, we propose a joint framework with a topic refinement mechanism to learn these two tasks simultaneously. Specifically, we design a three-pass iteration mechanism to generate coarse response first, then predict corresponding topics, and finally generate refined response conditioned on predicted topics. Moreover, we utilize GPT2DoubleHeads and BERT for the topic prediction task respectively, aiming to investigate the effects of joint learning and the understanding ability of GPT model. Experimental results demonstrate that our proposed framework achieves new state-of-the-art performance at response generation task and the great potential understanding capability of GPT model.

\end{abstract}

\section{Introduction}
Natural Language Generation (NLG), is the task of generating language that is coherent and understandable to humans, has been applied to many downstream tasks such as text summary \citep{zhang-etal-2019-pretraining, barhaim2020arguments, cho2020better, huang-etal-2020-achieved, gholipour-ghalandari-ifrim-2020-examining}, machine translation \citep{li2020shallowtodeep, baziotis2020language, cheng-etal-2020-advaug, zou-etal-2020-reinforced}, and dialogue response generation \citep{radford2019language, ijcai2018-643, tuan2019dykgchat, zhao-etal-2020-knowledge-grounded, liu2020meddg,wolf2019transfertransfo}.

\begin{figure*}[ht]
	\centering
	\footnotesize
	\begin{tikzpicture}
	\draw (-5,0) node[inner sep=0] {\includegraphics[width=0.58\columnwidth, height=5cm, trim={0cm 0cm 0cm 0cm}, clip]{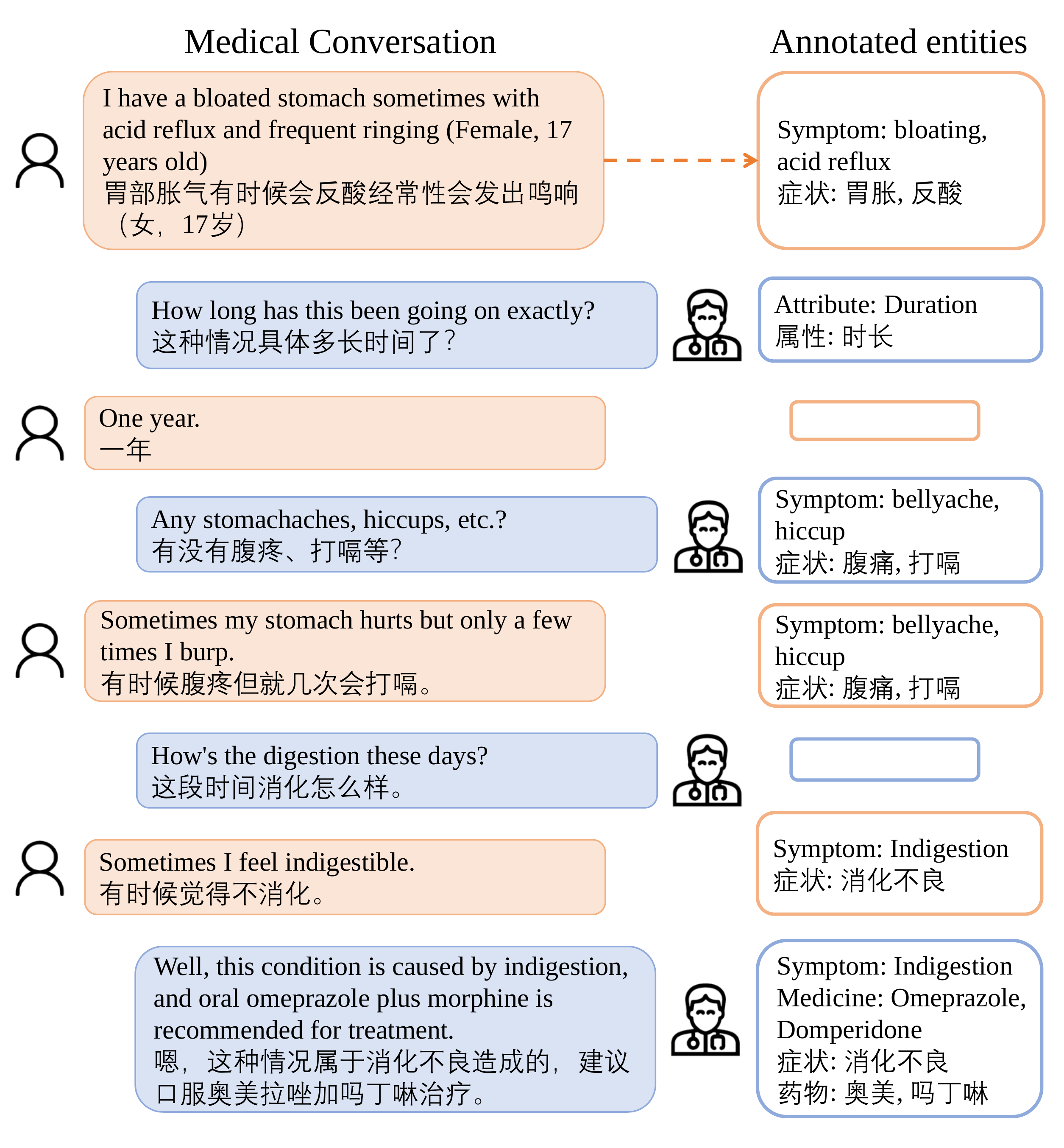}};
	
	\draw (3,0)  node[inner sep=0] {\includegraphics[width=1.2\columnwidth, height=4.2cm, trim={0cm 0cm 0cm 0cm}, clip]{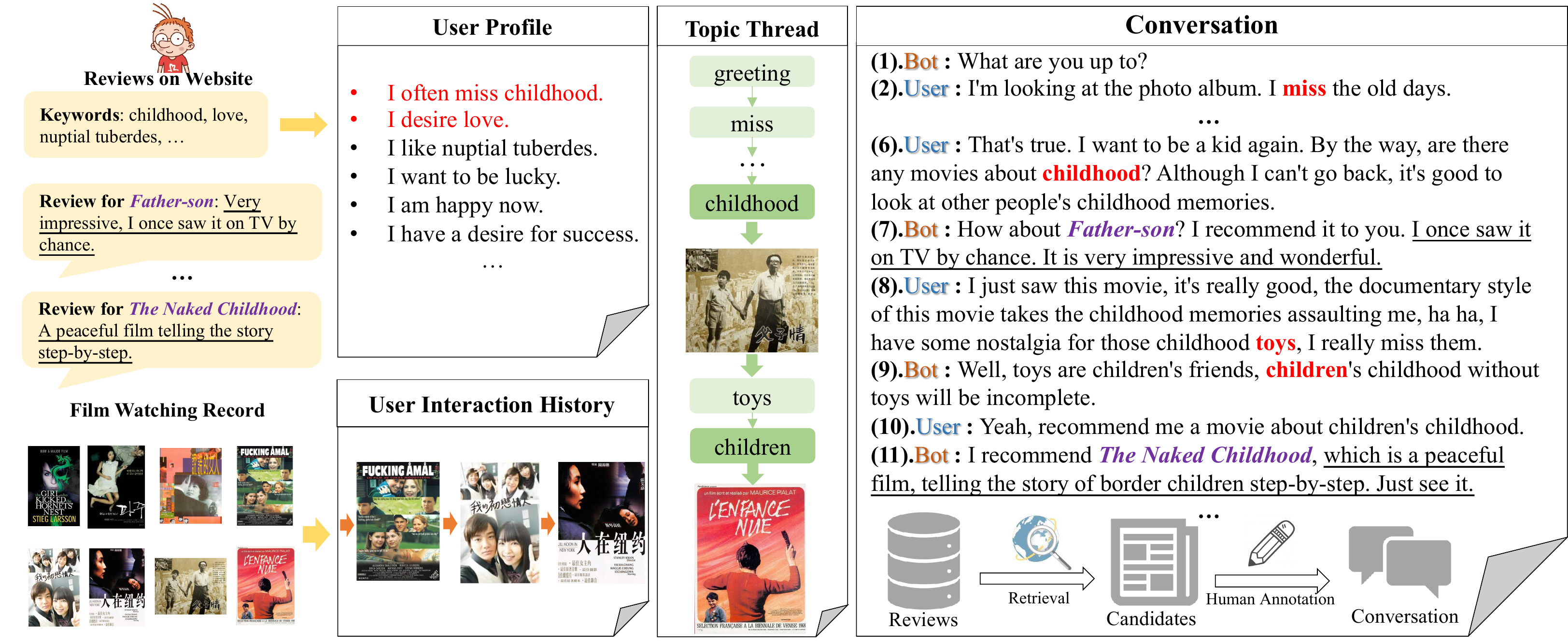}};
	\end{tikzpicture}
	\caption{\footnotesize
		\textbf{Left:} MedDG Dataset \textbf{Right:} TG-ReDial Dataset. Adapted from \cite{liu2020meddg} and \cite{liu-etal-2020-towards-conversational} respectively.
	}
	\label{fig:datasets}
\end{figure*}

Recent works in dialogue response generation usually formulate this task as a sequence to sequence problem, leading to inconsistent, uncontrollable, and repetitive responses \citep{ram2018conversational}. Furthermore, each dialogue has its specific goal and each utterance of the dialogue may contain multiple topics, regardless it is an open-domain dialogue or task-oriented dialogue. As shown in \textit{left} part of Figure~\ref{fig:datasets}, the patient seeks medical advice from a doctor and inform him the attributes and symptoms of the specific disease which form the topics of the conversation. Also, some open-domain dialogue systems have specific goals, such as recommendation, education, etc. For example, a conversational agent interacts with a user to recommend some interesting movies (as shown in \textit{right} part of Figure ~\ref{fig:datasets}). The entire content flow is guided by the topic thread. These various conversational scenarios propose more challenges for current multi-turn end-to-end dialogue system, necessitating the model's capability to generate more informative and topic-related response.

Many researchers propose different methods to guide or control the generation of responses conditioned on specific topics. Some representative works consider incorporating topic information into the sequence-to-sequence framework which apply attention mechanism to all topics \citep{xing2016topic, dziri2019augmenting}. Other works cast this task as a pipeline system, predict the keywords, then capture the topic, and finally retrieve corresponding response \citep{tang2019targetguided, zhou-etal-2020-towards}. Another line of work focuses on single-turn topic-aware response generation conditioned on previously given topics \citep{ijcai2018-0567, yang-etal-2019-enhancing, 9206719}. All these methods fall short in two ways. Most of these approaches either heavily rely on the non-autoregressive models like BERT \citep{devlin2019bert} to predict topics or utilize attention mechanism on all pre-defined topics which do not consider the effect of the historical topic path of multi-turn conversations. Besides that, these works do not model the joint distribution of attribute model $p(a|x)$ and unconditional language model $p(x)$, which is proved effective and powerful \citep{dathathri2019plug}.

% dialogue ground with some topic
In this paper, we formulate this problem as a topic-aware dialogue response generation task, aiming to generate informative and topic-related responses that can engage the users. More specifically, we design a three-stage iteration mechanism for the topic-aware response generation task. We generate the coarse response given historical dialogue context and previous topics first, then we require the model to explicitly predict corresponding topics, and then we concatenate the generated coarse response at the first step and the predicted topics at the second step as input to generate a final refined topic-related response. Thus, the model is forced to learn a joint distribution of topics and related response by optimizing for these three objectives simultaneously.

\begin{itemize}
    \setlength\itemsep{0pt}
    \item We formulate a traditional response generation problem as a topic-aware generation problem and propose a joint framework that can learn topic prediction and dialogue response generation simultaneously.
    \item We design a topic refine mechanism to control the generation of dialogue response. Our ablation study confirms it can help to generate more informative and topic-related responses, leading to better performance.
    \item We evaluate our model on two different datasets which consist of two application scenarios: medical auto-diagnosis and conversational recommendation, and we achieve new state-of-the-arts performance on both datasets and demonstrate that joint distribution and topic refinement are effective but the understanding ability of GPT2 still needs to be improved.

\end{itemize}

\section{Problem Definition}
Given a dialogue $ d = \{ u^1, u^2, u^3, ..., u^n \}$, a corresponding speaker role path $ sr = \{ s^1, s^2, s^3, ..., s^n \}$ and its corresponding topic path $tp = \{ tw^1, tw^2, tw^3, ..., tw^n \} $ where $ s \in R $, $tw \in T $. $R$ and $T$ are pre-defined speaker set and topic set. An utterance at $i$th time step can be expressed by $ (u^i, s^i, tw^i) $ which represents the sentence, the speaker and the topics included in this sentence. $tw^i$ consists of multiple topics or zero topic and each topic is expressed by several words. The problem then can be defined as: given a $i$th historical dialogue context, speaker role and topic path, $d_i^{n-1} = \{ u_i^1, u_i^2, ..., u_i^{n-1}\}, sr_i^{n-1} = \{s_i^1, s_i^2, ..., s_i^{n-1}\}, tp_i^{n-1} = \{tw_i^1, tw_i^2, ..., tw_i^{n-1}\}$, find the next topic and generate related responses.

\begin{equation}
    y^* = \operatorname*{arg\,max}_\theta p ( r^n, tw^n | d^{n-1}, tp^{n-1}, sr^{n-1})
\end{equation}

where $r^n$ and $tw^n$ stand for the response and corresponding topics at turn $n$ respectively,. User profile information  $p = \{ p_1, p_2, ..., p_k \}$ is often provided as additional input, which consists of $k$ sentences to express personal information such as interest. Thus, the objective changes accordingly:

\begin{equation}
    y^* = \operatorname*{arg\,max}_\theta p ( r^n, tw^n | d^{n-1}, tp^{n-1}, sr^{n-1}, p)
\end{equation}

Different from other methods, we divide the whole problem into three sub-problems (see section below). Our objective can be formulated as the following joint distribution:

\begin{equation}
\begin{split}
    y^* = \operatorname*{arg\,max}_\theta & p ( r_1^n | d^{n-1}, sr^{n-1}, tp^{n-1}) \\
        & p ( tw^{n} | d^{n-1}, sr^{n-1}, tp^{n-1} )  \\
        & p ( r_2^n | d^{n-1}, sr^{n-1}, tp^{n-1}, (r_1^n, tw^n) )
\end{split}
\end{equation}

where $p ( r_1^n | d^{n-1}, sr^{n-1}, tp^{n-1})$ generate the relatively abbreviated response first, then $p ( tw^{n} | d^{n-1}, sr^{n-1}, tp^{n-1} ) $ predict the corresponding topics at turn $n$, and finally the model refine the abbreviated response $r_1^n$ by maximizing $p ( r_2^n | d^{n-1}, sr^{n-1}, tp^{n-1}, (r_1^n, tw^n) )$ with the first response $r_1^n$ and corresponding predicted topics $tw^n$ as additional input, which leads to more informative and topic-related response $r_2^t$.

\section{Model}
\begin{figure*}[ht]
	\centering
	\footnotesize
	\begin{tikzpicture}
	\draw (0,0) node[inner sep=0] {\includegraphics[width=2.0\columnwidth, height=9cm, trim={0cm 2cm 0cm 0cm}, clip]{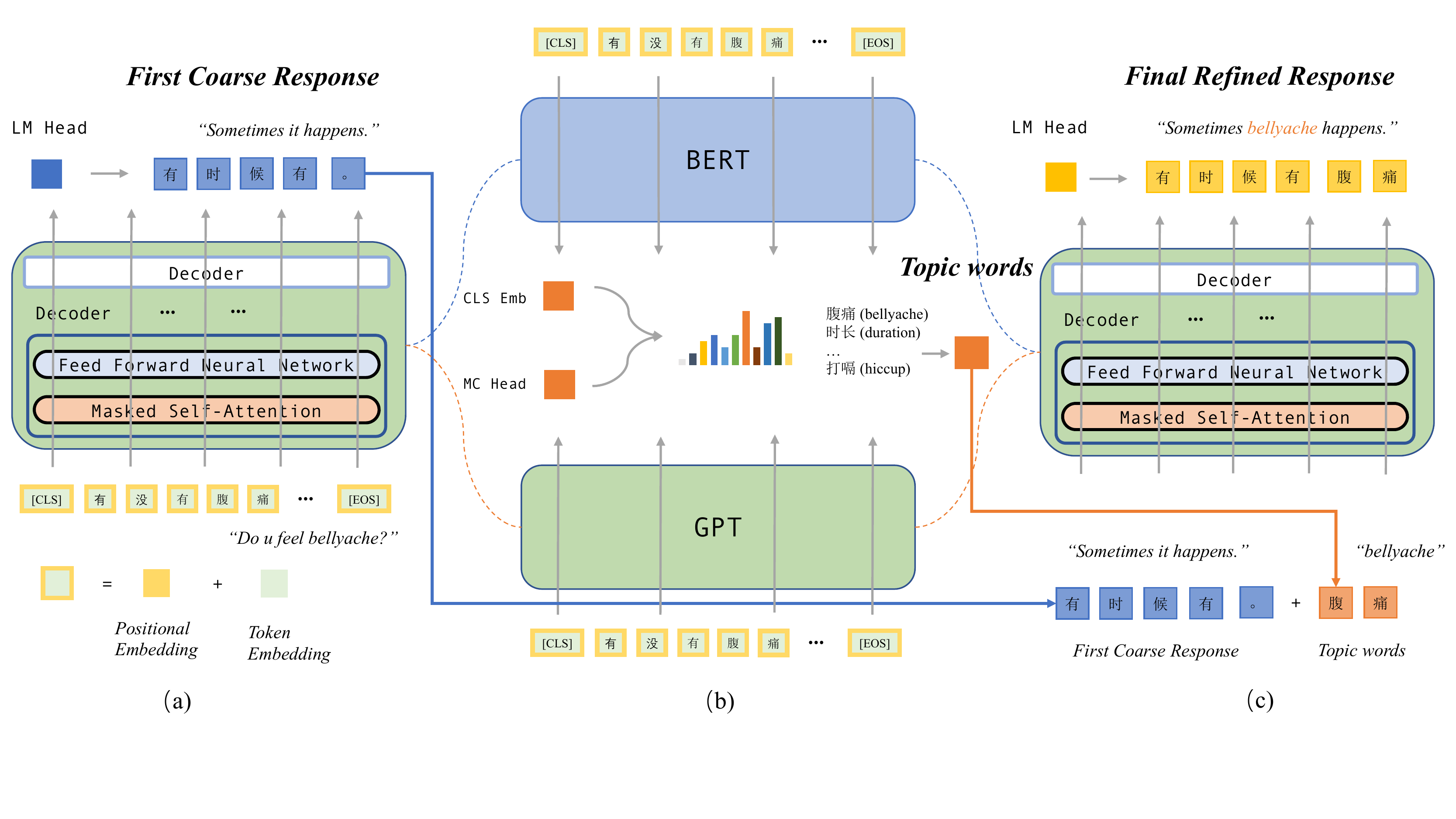}};
	\end{tikzpicture}
	\caption{\footnotesize
		TopicRefine: Joint Framework of Our Proposed Model, which consists of three different modules (a) Stage-One: Response Generation (b) Topic Prediction (c) Stage-Two: Topic Refinement. The (b) module can be implemented by two methods: BERT and GPT, we utilize Stage-One (GPT) and Stage-Two (GPT) to represents the framework with GPT as backbone for all three modules (\textcolor{orange}{orange dashed line}), and Stage-Two (BERT) to replace GPT with BERT for (b) module (\textcolor{blue}{blue dashed line}) in later experiment section. 
	}
	\label{fig:topicRefine}
\end{figure*}

Our model can be divided into three different parts: 1) Stage-One: Response Generation and 2) Topic Prediction; and 3) Stage-Two: Topic Refinement, which corresponds (a), (b), (c) shown in Figure~\ref{fig:topicRefine} respectively. More details can be checked in following subsections 3.1, 3.2, and 3.3.

\subsection{Stage-One: Response Generation}
% token type embeeding token embeeding 
% Few-shot Natural Language Generation for Task-Oriented Dialog section3
We formulate the response generation problem using conditional language models e.g. GPT \citep{radford2019language}. Given many dialogues $D = \{d_1, d_2, d_3, ..., d_m\}$, $i$th dialogue $d$ contains serval training samples $(r^n, tw^n | d^{n-1}, sr^{n-1}, tp^{n-1})$ from different turn $n$, our objective here is to build a statistical model parameterized by $\theta$ to maximize $p_{\theta} (r^n | d^{n-1}, tp^{n-1}, sr^{n-1})$. Since here we use autoregressive language models to take account of the sequential structure of the response, we need to decompose the joint probability of $r^n$ using the chain rule as follows:

\begin{equation}
    p_{\theta} (r^n | d^{n-1}, tp^{n-1}, sr^{n-1}) = \prod_{t=1}^{T} p_{\theta} (r_t^{n} | r_{<t}^{n}, d^{n-1}, tp^{n-1}, sr^{n-1})
\end{equation}

where $r_{<t}^{n}$ represents all tokens before $t$ at turn $n$. The objective of stage-one is performed by maximizing the loglikelihood (MLE) of the conditional probabilities in (4) over the entire training dataset:

\begin{equation}
    L_{one} = - \sum_{m=1}^{|D|} \sum_{n=1}^{|d|} \sum_{t=1}^{T} log p_{\theta} (r_t^{m,n} | r_{<t}^{m,n}, \mathcal{H}_m)
    \label{eq:one}
\end{equation}

where $r_{m,n}^t$ is $t$th token of $n$th resposne of $m$th dialogue in training dataset, $\mathcal{H}_m$ represents $ (d^{m,n}, tp^{m,n}, sr^{m,n} )$ before current response.

\subsection{Topic Prediction}
% data leak problem
Given the historical $\mathcal{H}_m$ of $m^{th}$ dialogue \footnote{It is noted that we do not use $r_{<t}^n$ as input information here.}, we need not only to generate the suitable response but also to predict the correct topic. Some prior works solve this problem by predicting the topic first and then generating the response \citep{liu2020meddg, zhou-etal-2020-towards}. In this section, different from these works, we propose a framework to jointly learn this task with dialogue response generation task as shown in Figure~\ref{fig:topicRefine}. There are two methods to predict the corresponding topics: (1) BERT-Based Prediction, and (2) GPT-Based Prediction.

\subsubsection{BERT-Based Prediction. } Consistent with previous work in text classification \citep{chen2019bert}, we utilize the embedding $h_1$ of first token $[CLS]$ from BERT \citep{devlin2019bert} to predict the topics, followed by a $softmax$ layer.

\begin{equation}
    f(x) = \mathrm{softmax} (Wh_1 + b)
\end{equation}

\subsubsection{GPT-Based Prediction. } We adapt GPT2DoubleHeads model \citep{wolf-etal-2020-transformers} to perform the prediction followed \cite{wolf2019transfertransfo}, since there are two heads: language modeling head and the classification head in the model while the later one can be used to classify the input dialogue information. Besides that, the shared parameters of GPT may benefit both topic prediction and response generation tasks. 

% understanding
% joint learning of same model
It is noted that there are two types of classification in topic prediction task: \textit{multi-class classification} and \textit{multi-label classification}, owing to the unique characteristic and differences of two dataset: MedDG \citep{liu2020meddg} and TG-ReDial \citep{liu-etal-2020-towards-conversational}. For a \textit{multi-class classification} problem, the global optimization can be reached by minimizing cross-entropy loss defined as follow:

\begin{equation}
    L_{topic} = - \sum_{c=1}^{K} y_c log (p_c | \mathcal{H}_m)
\end{equation}

For a \textit{multi-label classification} problem, it is usually formulated as a sequence of binary decision problems which are optimized by: 

\begin{equation}
    L_{topic} = - \sum_{c=1}^{K} y_c log (p_c | \mathcal{H}_m) + (1 - y_c) log(p_c | \mathcal{H}_m)
\end{equation}

\subsection{Stage-Two: Topic Refinement}
% topic contains multis section 2
To generate more informative and topic-related response, we introduce the \textit{topicRefine} mechanism that refines the generated response condition on the predicted topic \footnote{If there are k topics predicted by module b, then we simply concatenate all of them together}, as shown in Figure~\ref{fig:topicRefine} (c).

The refine decoder receives the first generated response $r_1^n$ from the stage-one module and the predicted topic $tw^n$ from the Topic Prediction module as input and outputs a refined response $r_2^n$. More specifically, we utilize $<topic>$ to indicate the position of topics, so the input can be represented as $\{ [CLS], w_r^1, w_r^2, ..., w_r^n, <topic>, w_t^1, w_t^2, ..., w_t^n, <topic> \}$ where $r_1^n = [w_r^1, w_r^2, ..., w_r^n]$, $tw^n = [w_t^1, w_t^2, ..., w_t^n]$. The learning objective is formulated as:

\begin{equation}
    L_{refine} = - \sum_{m=1}^{|D|} \sum_{n=1}^{|d|} \sum_{t=1}^{T} log p_{\theta} (r_t^{m,n} | r_{<t}^{m,n}, \mathcal{H}_m, tw^{n})
    \label{eq:refine}
\end{equation}

where Eq~\ref{eq:refine} is similar with Eq~\ref{eq:one} except the introduced topic information $tw^n$ here. The parameters are shared by all three modules unless we state otherwise.

\subsection{Training Objective}
The learning objective of our model is the sum of three parts, jointly trained using the ``teacher-forcing" algorithm. During training, we feed the ground-truth response only in stage-one and stage-two and minimize the following objective.

\begin{equation}
    L_{model} = L_{one} + L_{topic} + L_{refine}
\end{equation}

At test time, we choose the predicted word by $y^* = argmax_y p(y|x) $ at each time step, and we use greedy search to generate both the response and refined response.

\section{Experiment}
In this section, we will introduce datasets and baselines first, and then presents implementation details and evaluation metrics of our proposed framework.

\subsection{Datasets}

\noindent \textbf{MedDG} \citep{liu2020meddg} A large-scale high-quality medical dialogue dataset which contains 12 types of common diseases, more than 17k conversation, 160 different topics consisting of symptoms and attributes. Noted the topic-prediction task here is a multi-label classification problem.

\noindent \textbf{TG-ReDial} \citep{zhou-etal-2020-towards} consists of 10000 two-party dialogues between the user and a recommender in the movie domain which explicitly incorporates topic paths to enforce natural semantic transitions towards recommendation scenario. For topic-prediction task here, it is a multi-class classification problem. The details of these two datasets can be found in Table~\ref{tab:stat}.

\begin{table}[h]
\centering
\scalebox{0.9}{
\begin{tabular}{l|c|c}
Dataset & MedDG & TG-ReDial \\
\hline
Task Domain &  Task-oriented  & Recommendation \\
Language & Chinese & Chinese \\
Classification Type & Multi-Label & Multi-Class \\
Dialogue Domian & Medical & Movie \\
$\sharp$ Dialogues & 17864 & 10000 \\
$\sharp$ Utterances & 385951 & 129392 \\
$\sharp$ Topics & 160 & 2571 \\
\hline
\end{tabular}}
\caption{Statistics of Two Datasets}
\label{tab:stat}
\end{table}

\subsection{Baselines}

\noindent \textbf{Seq2Seq. } \citep{sutskever2014sequence} is a classical attention-based sequence to sequence model which build on top of vanilla RNN encoder and decoder. 

\noindent \textbf{HRED. } \citep{serban2016building} extends the traditional RNN encoder by stacking two RNNs in a hierarchical way: one at word level and one at the utterance level. It is frequently used as a dialogue encoder.

\noindent \textbf{GPT2. } \citep{radford2019language} is a strong baseline for response generation task which demonstrates powerful performance in many related works. It is noted all three methods mentioned above can utilize topic information as additional input which concatenate with utterance in the dialogue. We use \textbf{Seq2Seq-Topic}, \textbf{HRED-Topic} and \textbf{GPT-Topic} to represent these methods respectively.

\noindent \textbf{Redial} \citep{li2019deep} is proposed specially for conversational recommendation systems by utilizing an auto-encoder for recommendation.

\noindent \textbf{KBRD} \citep{chen2019knowledgebased} stands for Knowledge-Based Recommendaer Dialog System, which combines the advantages of recommendation system and dialogue generation system.

\noindent \textbf{Transformer} \citep{vaswani2017attention} applies a Transformer-based encoder-decoder framework to generate proper responses.

\noindent \textbf{TG-RG} \citep{zhou-etal-2020-towards} is current state-of-the-art method comes with the release of dataset. It predicts the topic first and then generates the response.

\subsection{Variants of Our Framework}

\noindent \textbf{GPT2DH. } The method removes the refinement stage from our framework and jointly train the response generation and topic prediction tasks (i.e. a and b module in Figure ~\ref{fig:topicRefine}) based on the GPT2DoubleHeads model. In this way, the training objective changes to $L_{model} = L_{one} + L_{topic}$ without $L_{refine}$. We called this method GPT2DH to represent GPT2DoubleHeads \citep{wolf-etal-2020-transformers} which have two heads for classification and generation respectively.

\noindent \textbf{Stage-One (GPT) and Stage-Two (GPT). } As shown in Figure~\ref{fig:topicRefine}, this variant represents all three components are implemented by GPT2DoubleHeads model, while \textbf{
Stage-One (GPT)} represents the first generated response $r_1^n$ and \textbf{Stage-Two (GPT)} represents the refined response $r_2^n$ in Equation (3).

\noindent \textbf{Stage-Two (BERT). } We replace GPT with BERT only for (b) module in Figure~\ref{fig:topicRefine}. The variant is designed for poor understanding capability of GPT model which leads to noisy predicted topic.

\subsection{Implementation Details}
We use the same settings for these two datasets. The learning rate is set as 1.5e-4, repetition penalty as 1.0, batch size as 4, warmup steps as 2000, except max context length as 500, max decode length as 50, epochs as 20 for TG-ReDial, max context length as 600, max decode length as 100, epochs as 10 for MedDG. We use \verb|ADAMW| \citep{loshchilov2019decoupled} to train the model. We emphasize that the role path information is missing in the test data of MedDG. Thus we only use dialogue and topic information in the experiment to keep consistent with test data. It is important to note that our methods do not pre-train on any other big corpus, we just load the parameters provided by \cite{wolf-etal-2020-transformers} and directly fine-tune on the target dataset.

\subsection{Evaluation Metrics}

For the sake of fair comparison, we adopt the same evaluation metrics as the original two papers \citep{liu2020meddg} and \citep{zhou-etal-2020-towards}. For MedDG, we report BLEU-1, BLEU-4, and Topic-F1 for response generation task, and Precision, Recall, and F1 score for the topic prediction task. For TG-ReDial, we calculate BLEU-1, BLEU2, and BLEU3 for generation and Hit@1, Hit@3, Hit@5 for prediction. It is noted that Topic-F1 requires the topic words appears exactly in the generated response at MedDG dataset.

\section{Result and Analysis}

In this section, we evaluated the proposed TopicRefine framework at two datasets MedDG and TG-ReDial respectively. And then we further investigate the effects of different response length and provide analysis of human evaluation for dialogue response generation task. At the last, we also investigate the understanding capability of GPT model at these two datasets.

\subsection{Main Result}

\begin{table}[h]
\centering
\scalebox{0.8}{%
\begin{tabular}{l|cccc}
\toprule[1pt]
Model  & BLEU-1 & BLEU-4 & Topic-F1 & Avg Score\\
\hline
Seq2Seq & 26.12 & 14.21 & 12.63 & 17.65 \\
Seq2Seq-Topic & 35.24 & 19.20 & 16.73 & 23.72\\
HRED & 31.56 & 17.28 & 12.18 & 20.34\\
HRED-Topic & 38.66 & 21.19 & 16.58 & 25.48\\
GPT2 & 29.35 & 14.47 & 9.17 & 17.66\\
GPT2-Topic & 30.87 & 16.56 & 17.08 & 21.50 \\
\hline
Stage-Two (GPT) & \textbf{45.12} & \textbf{27.49} & 5.40 & 26.00\\
Stage-Two (BERT) & 44.49 & 24.62 & \textbf{17.94} & \textbf{29.02} \\
\hline
Stage-One (GPT) &	43.86 &	24.62 & 11.36 & 26.61 \\
GPT2DH & 43.93 & 24.35 & 11.91 & 26.73 \\
\bottomrule[1pt]
\end{tabular}}
\caption{Dialogue response generation at MedDG dataset. It is notes that ``-Topic" methods use the ground truth topic information in the dataset.}
\label{tab:meddg_gen_exp}
\end{table}

\begin{table}[h]
\centering
\scalebox{0.9}{%
\begin{tabular}{l|lll}
\toprule[1pt]
Model  & BLEU-1 & BLEU-2 & BLEU-3 \\
\hline
Redial & 0.177 & 0.028 &  0.006 \\
KBRD & 0.223 & 0.028 & 0.009 \\
Transformer & 0.283 & 0.068 & 0.033 \\
GPT2-Topic & 0.278 & 0.064 & 0.031 \\
TG-RG & 0.282 & 0.067 & 0.033 \\
\hline
Stage-Two (GPT) & 0.293 & 0.085 & 0.042 \\
Stage-Two (BERT) & \textbf{0.294} & \textbf{0.086} & \textbf{0.043}\\
\hline
Stage-One (GPT) & 0.284 & 0.082 & 0.041 \\
GPT2DH & 0.288 & 0.086 & 0.041 \\
\bottomrule[1pt]
\end{tabular}}
\caption{Recommendation Response Generation at TG-ReDial dataset.  It is notes that ``-Topic" methods use the ground truth topic information in the dataset.}
\label{tab:tg_gen_exp}
\end{table}

\begin{CJK*}{UTF8}{gbsn}
\begin{table*}[t]
\centering
\scalebox{0.6}{%
\begin{tabular}{l||p{10cm}|p{14cm}}
\hline
Dataset & MedDG & TG-ReDial  \\
\hline
\multirow{3}*{GroundTruth} & 可以服用多潘立酮改善一下症状。(You can take domperidone to alleviate.) Topics: [``多潘立酮"
] [``domperidone"] & 还真不是很了解，感觉不就是黑白画嘛，我给你推荐个黑白电影看看吧。 (I really don’t know it very well. I think it’s just "black-and-white" painting. Let me recommend a "black-and-white" movie for you.) Topics: [``黑白电影"] [``black-and-white movie"] \\
& 别客气有不懂的再及时沟通祝康复！ (Never mind, if you are still confused, talk with me and best wishes to you!) Topics: [None] & 那我再给你推荐个关于回忆的电影怎么样？(How about I recommend another movie about memories?) Topics: [``回忆"] [``memories"]\\
& 我建议你去买兰索拉唑，枸橼酸铋钾，阿莫西林，如果还是拉肚子的话，买一两包蒙脱石散，一次一包，不拉肚子就不用在吃了。 (I suggest you buy lansoprazole, bismuth potassium citrate, amoxicillin, if you still have diarrhea, buy one or two packs of montmorillonite powder, one pack at one time, don’t need to eat if you don’t have diarrhea.) Topics: ["腹泻", "阿莫西林", "蒙脱石散", "兰索拉唑"] ["diarrhea", "amoxicillin", "montmorillonite powder", "lansoprazole"] & 是啊，寂寞的时候可以看看。 (Yes, you can watch it when you are lonely.) Topics: [``亲情"] [``family"] \\
\hline

\multirow{3}*{Stage-One}  & 嗯嗯，祝你早日康复。(ok, I wish you can recovery soon.) & 我也喜欢，我给你推荐一部黑白电影吧。《影人》，良心推荐你去看看，黑白色电影，画面非常美。但是美国人物 (I like it too. Let me recommend a "black-and-white" movie to you. "Shadowman", truly recommends you to see it, "black-and-white" movie, the picture is very beautiful. But American characters) \\
& 除了胃烧心，反酸，别的没有。 (Except for heartburn and acid reflux, nothing else.) & 是啊，我们都是这样的回忆，我再给你推荐个关于回忆的电影吧！你看过《我回忆》吗，这个电影怎么样，这个电 (Yes, we all experienced these, let me recommend you a movie about memories! Have you watched "My Memories", how about this movie, this movie) \\
& 你自己什么时候开始疼的。(When did you start to feel hurt.) & 那我给你推荐一部有关英雄的亲情电影吧，《小情》挺好的，这部电影是我看过的最爽的一部。。超越亲情的一部 (Then I recommend you a family movie about heroes. "Little Love" is very good. This movie is the coolest one I have ever seen. . A piece beyond family) \\
\hline
\multirow{3}*{Stage-Two} & 幽门螺杆菌感染。根据病因治疗。 (Helicobacter pylori infection. Treat according to the cause.) & 我也喜欢，我给你推荐一部关于黑白的电影吧。你可以看看。《午夜夜幕》，黑白色调，还是很有趣的。 (I like it too. Let me recommend you a movie about black and white. You can have a look. "Midnight Curtain", black and white tones, is still very interesting.) \\
& 考虑受凉。可以服用雷贝拉唑搭配多久一次一粒，一天两次，餐前半小时服用 (Consider catching a cold. How often can you take rabeprazole with one pill, twice a day, half an hour before a meal) & 是啊，我们都是这样的回忆啊。我再给你推荐个关于回忆的电影吧！你看过《我们的童年》吗？这个电影很有意思 (Yes, we all have such memories. Let me recommend you another movie about memories! Have you watched "Our Childhood"? This movie is very interesting) \\
& 如果现在症状比较重，可以口服阿莫西林胶囊，奥美拉唑和小柴胡颗粒 (If the symptoms are severe now, you can take amoxicillin capsules, omeprazole and Xiaochaihu granules by mouth) & 那我给你推荐一部有关英雄的亲情电影吧，《秘罗》非常适合你，这部电影是我看过的最爱的英雄电影之一。  (Then I recommend you a family movie about heroes. "Mira" is very suitable for you. This movie is one of the favorite hero movies I have ever seen.) \\
\hline
\end{tabular}}
\caption{Generated Response Samples with Corresponding Topics for MedDG and TG-ReDial respectively. \textit{Stage-One} and \textit{Stage-Two} represent \textit{Stage-One (GPT)} and \textit{Stage-Two (GPT)} respectively.}
\label{tab:generation_sample}
\end{table*}
\end{CJK*}

Table~\ref{tab:meddg_gen_exp} and Table~\ref{tab:tg_gen_exp} demonstrates the performance of baselines and our proposed framework in both MedDG and TG-ReDial dataset respectively. Our topicRefine framework outperforms the previous state-of-the-art models at both datasets (i.e. GPT2-Topic model at MedDG and TG-RG model at TG-ReDial). More specifically, Stage-Two (GPT) reaches better BLEU score and Stage-Two (BERT) achieves higher Topic-F1 score at MedDG, owing to the existence of noisy topic in former method. Consistent with MedDG dataset, our method gets better performance no matter in Stage-Two (GPT) or Stage-Two (BERT) as shown in Table~\ref{tab:tg_gen_exp}. BLEU-1, BLEU-2, and BLEU-3 all have been improved by different degrees. Another interesting finding is that when explicitly concatenating topic words with dialogue utterances, the GPT-Topic model achieves a higher topic-f1 score, whereas the Stage-Two (GPT) model achieves a lower topic-f1 score, indicating the effectiveness of simply concatenating topic words and the noisy prediction results by GPT.

\subsection{Ablation Study}

To further investigate the effectiveness of our proposed framework, we add some variants of our proposed framework (i.e. Stage-One (GPT) and GPT2DH) as ablation study. As shown in Table~\ref{tab:meddg_gen_exp} and Table~\ref{tab:tg_gen_exp}, Stage-One (GPT) and GPT2DH achieve comparable results. On the one hand, compared with previous state-of-the-art models, GPT2DH demonstrate more powerful capability which shows the effectiveness of joint learning by incorporating topic prediction. Besides, any Stage-Two model reaches higher BLEU scores than GPT2DH which demonstrate the effectiveness of refine mechanism (i.e. $L_{refine}$). On the other hand, Stage-Two (GPT) outperforms Stage-One (GPT) in BLEU score (45.12 vs 43.86) but Topic-F1 score (5.40 vs 11.36). We argue that the model tends to generate more topic-related words instead of a specific topic word in the response. This is reasonable since the model is optimized to generate a more informative and topic-related response rather than a specific word.

\subsection{Effects of Response Length}

\begin{figure}[h]
\centering
\includegraphics[scale=0.35]{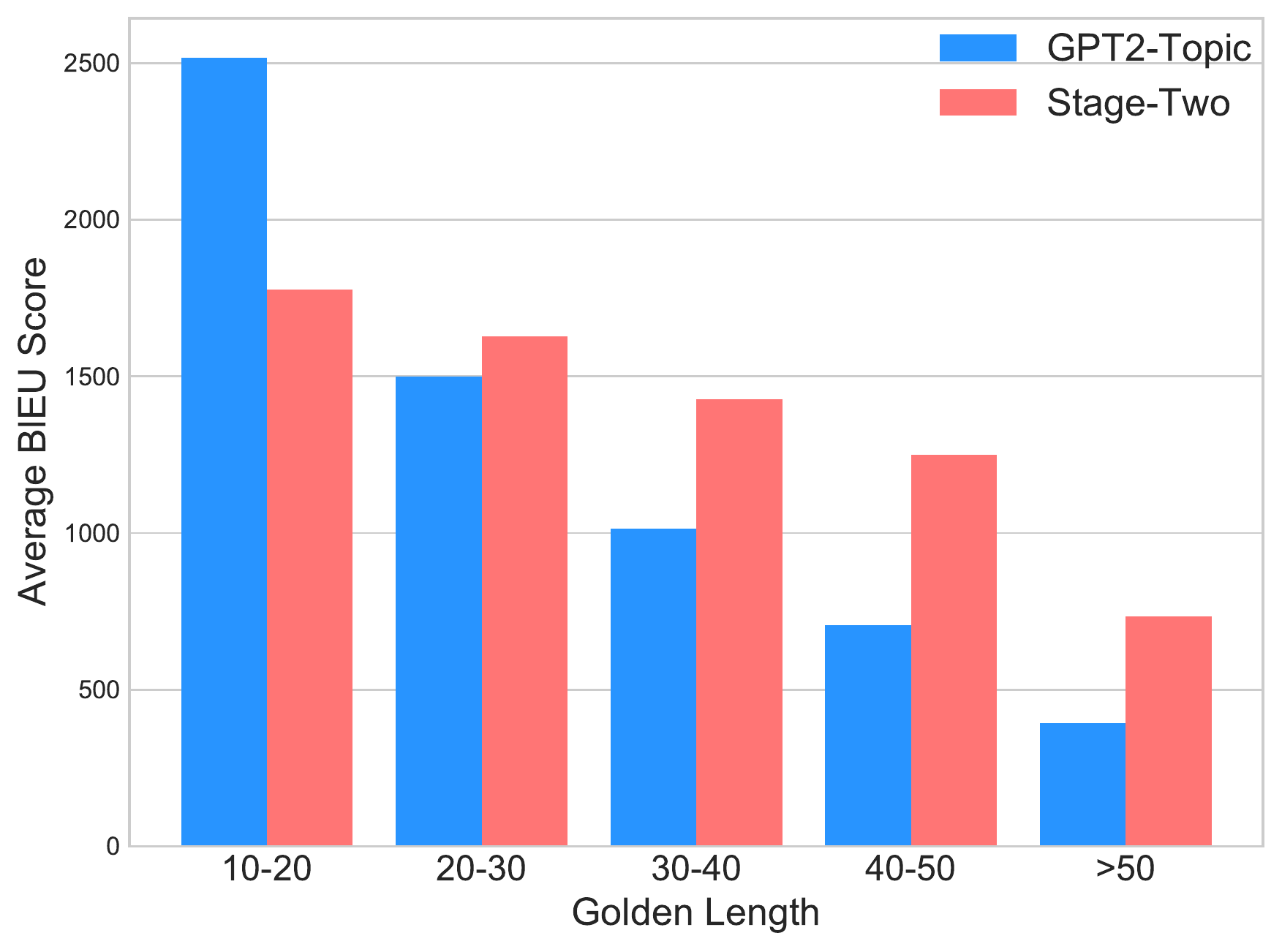}
\caption{Average BLEU score of MedDG for different golden length}
\label{fig:meddg_bleu}
\end{figure}

\begin{figure}[h]
\centering
\includegraphics[scale=0.35]{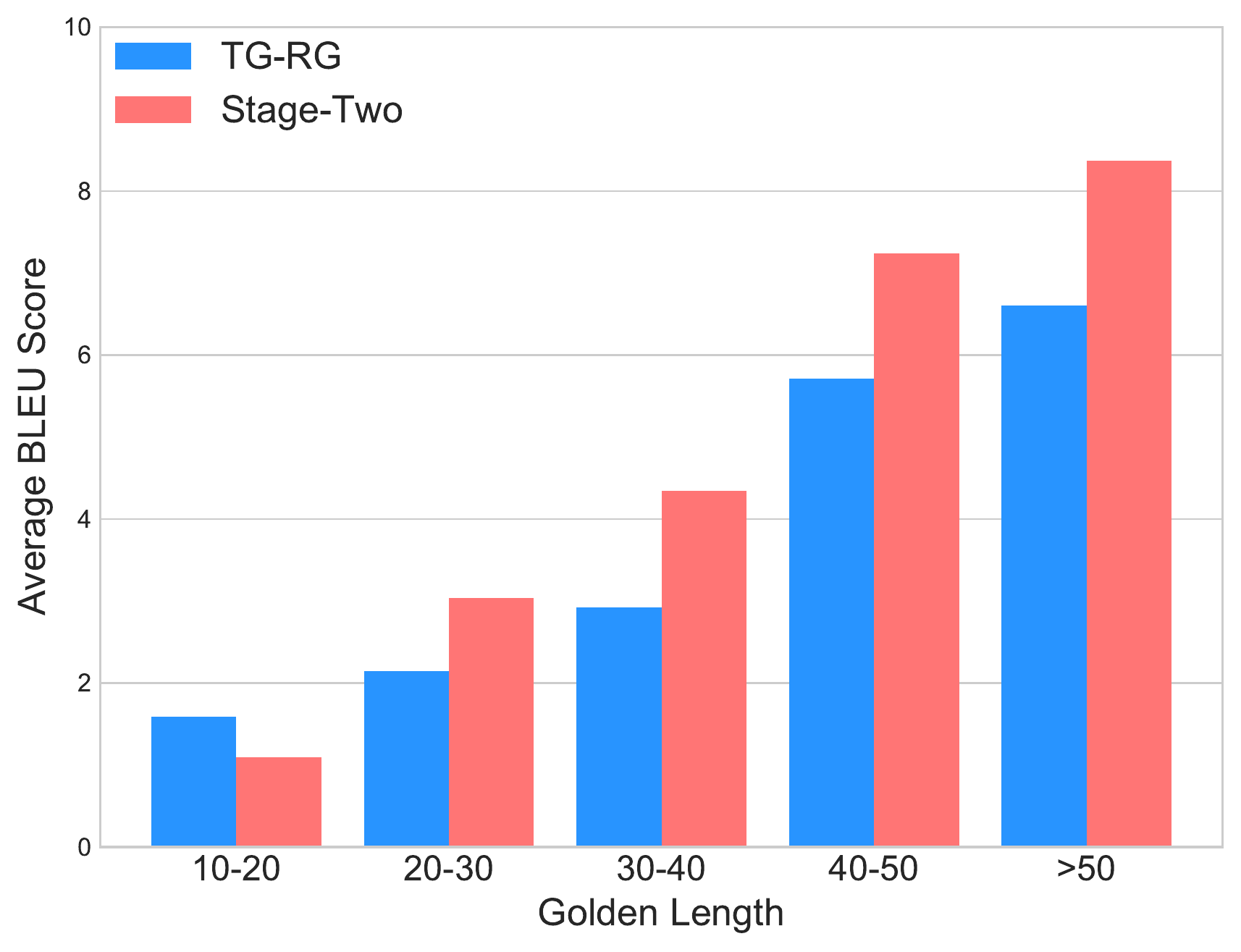}
\caption{Average BLEU score of TG-ReDial for different golden length}
\label{fig:tg_bleu}
\end{figure}

To evaluate the impact of different ground-truth response length, we compare the average BLEU score between our model and previous state-of-the-art model (i.e. GPT2-Topic and TG-RG) in MedDG and TG-ReDial respectively. As shown in Figure~\ref{fig:meddg_bleu} and Figure~\ref{fig:tg_bleu}, our model reaches better performance when the length of golden response is greater than 20 (occupies about 47.6\% and 81.9\% of test set respectively). As the golden length increases, our improvements also get boosted, which is more obvious at TG-ReDial dataset.

\subsection{Generated Sample}

Table~\ref{tab:generation_sample} given some generated response at both datasets. To summarize, our generated result has the following features:

\begin{CJK*}{UTF8}{gbsn}
\begin{itemize}
    \setlength\itemsep{0pt}
    \item For MedDG, since we drop the information of speaker role path during training and the dialogue between the doctor and the patient is not alternate, some generated responses may represent the perspective of the patient. 
    \item For TG-ReDial, there are some meaningless repeated characters in the result of Stage-One. For example, ``。" and ``这个电" (this movie) appears twice in response generated by Stage-One. Stage-Two can alleviate this problem by incorporating topic refinement.
    \item Our Stage-Two model can generates more informative response conditioned on given topics. Taking the sample of TG-ReDial in Table~\ref{tab:generation_sample} as an example. For the topic of ``memories", the response of ground truth is just a rhetorical question, while the response of our model not only grasps this topic, but also recommends one specific movie name related to this topic, which suggests that our model is able to ground multi-turn dialogue generation to some specific topics and tends to be more informative with respect to context.
\end{itemize}
\end{CJK*}

\subsection{Human Evaluation}

To perform human evaluation, we randomly select 50 examples from the outputs of previous sota model, and our \textit{Stage-One (GPT)} and \textit{State-Two (GPT)} method. The annotators are required to assign two scores for each sentence according to two criteria: (1) information and (2) fluency, ranging from 0 to 10. \textit{information} measures which sentence contains more information (e.g. less repetition). \textit{Fluency} measures which sentence is more proper as a response to a given dialogue context. The evaluation results are calculated by average these two scores of all sentences. 

\begin{table}[]
    \centering
    \begin{tabular}{l|cc|cc}
    \toprule[1pt]
     \multirow{2}{*}{Model} &  \multicolumn{2}{c}{MedDG} &  \multicolumn{2}{c}{TG-ReDial} \\ \cline{2-5} & Information & Fluency &  Information & Fluency \\
     \hline
      Human & 6.99 & 6.28 & 7.40 & 7.28 \\ 
      Baseline & 6.18 & 5.51 & 6.20 & 5.69 \\
      One & 6.32 & 4.81 & 6.62 & 5.66 \\
      Two & 6.57 & 6.13 & 7.30 & 6.42 \\
    \bottomrule[1pt]
    \end{tabular}
    \caption{The result of human evaluation. The baseline represents previous sota model \textit{GPT2-Topic} and \textit{TG-RG} in MedDG and TG-ReDial dataset respectively. One represents \textit{Stage-One (GPT)} and Two represents \textit{Stage-Two (GPT)} }
    \label{tab:human_eva}
\end{table}

Table~\ref{tab:human_eva} demonstrates the result of human evaluation. Generally, the score at TG-ReDial dataset is relatively higher than score in MedDG dataset. We attribute this to the MedDG dataset necessitates more expert knowledge and contains many terminologies. Besides that, there is still a large gap between generated response and human response, especially at fluency criteria. In detail, the Stage-One (GPT) perform better than baseline models at information but worse at fluency. Stage-Two (GPT) model gets better scores in both information and fluency criteria than Stage-One (GPT) model and baseline.

\subsection{Understanding of GPT Model}

\begin{table}[h]
\centering
\begin{tabular}{l||lll}
\hline
Model & P & R & F1 \\
\hline\
BERT & 14.48 & \textbf{32.95} & 20.13 \\
\hline
Stage-Two (GPT) & \textbf{22.22} & 11.16 & 14.88 \\
\hline
\end{tabular}
\caption{Result of topic prediction task (multi-label classification) at MedDG dataset}
\label{tab:meedg_und_exp}
\end{table}

\begin{table}[h]
\centering
\scalebox{0.9}{%
\begin{tabular}{l||lll}
\hline
Model & Hit@1 & Hit@3 & Hit@5 \\
\hline
BERT & \textbf{0.7651} & \textbf{0.8023} & \textbf{0.8189} \\
\hline
Stage-Two (GPT) & 0.5640 & 0.7931 & 0.8122 \\
\hline
\end{tabular}}
\caption{Result of topic prediction task (multi-class classification) at TG-ReDial dataset}
\label{tab:tg_und_exp}
\end{table}

Table~\ref{tab:meedg_und_exp} and Table~\ref{tab:tg_und_exp} demonstrate the performance of topic prediction task at MedDG and TG-ReDial datasets respectively. It is obvious that BERT \citep{devlin2019bert} demonstrates more strong understanding ability than GPT \citep{wolf-etal-2020-transformers} model. However, the comparable performance of Hit@3 and Hit@5 between BERT and GPT in Table~\ref{tab:tg_und_exp} clearly demonstrates the latter's high understanding potential. The unlocking of potential necessitates a more meticulously designed algorithm or architecture \citep{dathathri2019plug,liu2021gpt}.

\section{Related Work}

\subsection{Topic-aware Dialogue System}

% with external knowledge
Data-driven, knowledge-grounded dialogue system \citep{ijcai2018-643, tuan2019dykgchat, zhao-etal-2020-knowledge-grounded} attracts much attention due to the release of large pre-trained language models such as GPT2 \citep{radford2019language} and DialoGPT \citep{zhang2019dialogpt}. According to different types of knowledge, previous works can be clustered into the following categories: (1) attributes \citep{ zhou2018emotional, zhang-etal-2018-learning-control, xu-etal-2019-neural} (2) persona \citep{li-etal-2016-persona, zheng2019pretraining, wu2020guiding, zhang-etal-2018-personalizing} (3) external knowledge graph such as commonsense knowledge \citep{tuan2019dykgchat, yang-etal-2019-enhancing, moon-etal-2019-opendialkg}. 

Most of previous works for topic-aware dialogue system \citep{xing2016topic, dziri2019augmenting, yang-etal-2019-enhancing, 9206719} utilize attention mechanism on all topics at the decode stage to bias the generation probability. \cite{tang2019targetguided} proposes a structured approach that introduces coarse-grained keywords to control the intended content of system responses and \cite{xu2020topic} proposes Topic-Aware Dual-attention Matching (TADAM) Network to select suitable response but all of their systems are retrieval-based.

\subsection{Refine Mechanism}

Refine mechanism has been proved to be a effective and compelling technique in both natural language understanding and generation tasks \citep{zhang-etal-2019-pretraining, wu-etal-2020-slotrefine,song-etal-2021-bob}. For natural language understanding, \cite{wu-etal-2020-slotrefine} design a novel two-pass iteration mechanism to handle the uncoordinated slots problem caused by conditional independence of non-autoregressive model, in which the model utilizes \textit{B-label} output from first phase as input at second phase. For natural language generation, \cite{zhang-etal-2019-pretraining} use refine mechanism to generate refined summary which firstly applies BERT as decoder. Recently, a novel BERT-over-BERT (BoB) model is proposed to solve response generation task and consistency understanding simultaneously \citep{song-etal-2021-bob}. In this paper, we utilize \textit{topicRefine} framwork to build a topic-aware multi-turn end-to-end dialogue system, aiming to generate informative and topic-related dialogue response.

\begin{comment}
In this paper, we formulate this problem as a topic-aware dialogue response generation task and aim to generate topic-related informative and interesting responses that can engage the users. Inspired by recent investigation on joint distribution of attribute model and language model \cite{dathathri2019plug}, refinement mechanism \cite{zhang-etal-2019-pretraining, wu-etal-2020-slotrefine} and also the understanding capability of GPT \cite{liu2021gpt}, we propose to jointly learn topic prediction and response generation based on GPT2DoubleHead \cite{radford2019language, wolf-etal-2020-transformers}, and aim to 1) explore its natural language understanding and generation ability simultaneously, and 2) validate the effectiveness of joint distribution and refinement mechanism. To summarize, the technical contributions in this work are as follows:
\end{comment}

\section{Conclusion and Future Work}
In this paper, we propose a joint framework with a topic refinement mechanism to solve topic-aware multi-turn end-to-end dialogue generation problem based on the auto-regressive language model -- GPT2 \citep{wolf-etal-2020-transformers}. More specifically, we design a three-pass mechanism to jointly learn topic predication and dialogue response generation tasks, aiming to generate informative and topic-related response to engage user. Comprehensive experiments demonstrate that our method outperforms previous state-of-the-art model on both MedDG \citep{liu2020meddg} and TG-ReDial \citep{liu-etal-2020-towards-conversational} datasets, which verifies that the effectiveness of joint learning and refinement mechanism. We will investigate more refine techniques in our future work.

\bibliographystyle{aaai}
\bibliography{ref.bib}

\end{document}